\documentclass[twoside]{article}

\usepackage{aistats2020_modified}

\usepackage{graphicx}  
\usepackage{ amssymb }
\usepackage{amsmath}
\usepackage{bbm}
\usepackage{mathtools}
\usepackage{xcolor}

\usepackage{amssymb}
\usepackage{amsmath}
\usepackage{bbm}
\usepackage{mathtools}

\newcommand{\kth}{$k^\mathrm{th}$ }

\usepackage[numbers]{natbib}
\renewcommand{\bibname}{References}
\renewcommand{\bibsection}{\subsubsection*{\bibname}}

\begin{document}

	\twocolumn[
	
	\aistatstitle{Emergent properties of the local geometry of neural loss landscapes}
	
	\aistatsauthor{ 
		Stanislav Fort
		\And  
		Surya Ganguli }
	
	\aistatsaddress{ 
		Stanford University \\ 
		Stanford, CA, USA 
		\And  
		Stanford University \\ 
		Stanford, CA, USA  
	} 

]

	\begin{abstract}
	
	    The local geometry of high dimensional neural network loss landscapes can both challenge our cherished theoretical intuitions as well as dramatically impact the practical success of neural network training.  Indeed recent works have observed 4 striking local properties of neural loss landscapes on classification tasks: (1) the landscape exhibits exactly $C$ directions of high positive curvature, where $C$ is the number of classes; (2) gradient directions are largely confined to this extremely low dimensional subspace of positive Hessian curvature, leaving the vast majority of directions in weight space unexplored; (3) gradient descent transiently explores intermediate regions of higher positive curvature before eventually finding flatter minima; (4) training can be successful even when confined to low dimensional {\it random} affine hyperplanes, as long as these hyperplanes intersect a Goldilocks zone of higher than average curvature.  We develop a simple theoretical model of gradients and Hessians, justified by numerical experiments on architectures and datasets used in practice, that {\it simultaneously} accounts for all $4$ of these surprising and seemingly unrelated properties. Our unified model provides conceptual insights into the emergence of these properties and makes connections with diverse topics in neural networks, random matrix theory, and spin glasses, including the neural tangent kernel, BBP phase transitions, and Derrida's random energy model.

	\end{abstract}
	
	\section{Introduction}
	\label{sec:intro}
	
	The geometry of neural network loss landscapes and the implications of this geometry for both optimization and generalization have been subjects of intense interest in many works, ranging from studies on the lack of local minima at significantly higher loss than that of the global minimum \cite{Saxe2014-ch,Dauphin2014-lk} to studies debating relations between the curvature of local minima and their generalization properties \cite{Hochreiter1997-pc,Keskar2016-zh,Dinh2017-us,Chaudhari2016-mz}.   Fundamentally, the neural network loss landscape is a scalar loss function over a very high $D$ dimensional parameter space that could depend {\it a priori} in highly nontrivial ways on the very structure of real-world data itself as well as intricate properties of the neural network architecture.  Moreover, the regions of this loss landscape explored by gradient descent could themselves have highly atypical geometric properties relative to randomly chosen points in the landscape.  Thus understanding the shape of loss functions over high dimensional spaces with potentially intricate dependencies on both data and architecture, as well as biases in regions explored by gradient descent, remains a significant challenge in deep learning.  Indeed many recent studies explore extremely intriguing properties of the {\it local} geometry of these loss landscapes, as characterized by the gradient and Hessian of the loss landscape, both at minima found by gradient descent, and along the journey to these minima.  
	
	In this work we focus on providing a simple, unified explanation of 4 seemingly unrelated yet highly intriguing local properties of the loss landscape on classification tasks that have appeared in the recent literature:  
	
	\paragraph{(1) The Hessian eigenspectrum is composed of a bulk plus $C$ outlier eigenvalues where $C$ is the number of classes.} Recent works have observed this phenomenon in small networks \cite{sagun2016eigenvalues,sagun2017empirical,yao2018hessianbased}, as well as large networks \cite{papyan2019measurements,ghorbani2019investigation}. This implies that locally the loss landscape has $C$ highly curved directions, while it is much flatter in the vastly larger number of $D-C$ directions in weight space.  
	
	\paragraph{(2) Gradient aligns with this tiny Hessian subspace.}  Recent work \cite{gurari2018gradient} demonstrated that the gradient $\vec{g}$ over training time lies primarily in the subspace spanned by the top few largest eigenvalues of the Hessian $H$ (equal to the number of classes $C$). This implies that most of the descent directions lie along extremely low dimensional subspaces of high local positive curvature; exploration in the vastly larger number of $D-C$ available directions in parameter space over training utilizes a small portion of the gradient.  
	
	\paragraph{(3) The maximal Hessian eigenvalue grows, peaks and then declines during training.}
		 Given widespread interest in arriving at flat minima (e.g. \cite{Chaudhari2016-mz}) due to their presumed superior generalization properties, it is interesting to understand how the local geometry, and especially curvature, of the loss landscape varies over training time.  Interestingly, a recent study \cite{jastrzebski2018relation} found that the spectral norm of the Hessian, as measured by the top Hessian eigenvalue, displays a non-monotonic dependence on training time. This non-monotonicity implies that gradient descent trajectories tend to enter higher positive curvature regions of the loss landscape before eventually finding the desired flatter regions. Related effects were observed in \cite{keskar2016large,NIPS1992_589}.
	
    \paragraph{(4) Initializing in a Goldilocks zone of higher convexity enables training in very low dimensional weight subspaces.}
		Recent work \cite{li2018measuring} showed, surprisingly, that one need not train all $D$ parameters independently to achieve good training and test error; instead one can choose to train {\it only} within a {\it random} low dimensional affine hyperplane of parameters. Indeed the dimension of this hyperplane can be {\it far less} than $D$.  More recent work \cite{fort2018goldilocks} explored how the success of training depends on the position of this hyperplane in parameter space.  This work found a Goldilocks zone as a function of the initial weight variance, such that the intersection of the hyperplane with this Goldilocks zone correlated with training success.  Furthermore, this Goldilocks zone was characterized as a region of higher than usual positive curvature as measured by the Hessian statistic $\mathrm{Trace}(H)/||H||$. This statistic takes larger positive values when typical randomly chosen directions in parameter space exhibit more positive curvature \cite{fort2018goldilocks,fort2019large}.  Thus overall, the ease of optimizing over low dimensional hyperplanes correlates with intersections of this hyperplane with regions of higher positive curvature.  
		
    Taken together these $4$ somewhat surprising and seemly unrelated local geometric properties fundamentally challenge our conceptual understanding of the shape of neural network loss landscapes. It is {\it a priori} unclear how these $4$ properties may emerge naturally from the very structure of real-world data, complex neural architectures, and potentially biased explorations of the loss landscape through the dynamics of gradient descent starting at a random initialization.   Moreover, it is unclear what {\it specific} aspects of data, architecture and descent trajectory are important for generating these $4$ properties, and what myriad aspects are not relevant. 
    
    Our main contribution is to provide an extremely simple, unified model that simultaneously accounts for {\it all} 4 of these local geometric properties.  Our model yields conceptual insights into why these $4$ properties might arise quite generically in neural networks solving classification tasks, and makes connections to diverse topics in neural networks, random matrix theory, and spin glasses, including the neural tangent kernel \cite{jacot2018neural,lee2019wide}, BBP phase transitions \cite{Baik2005-rh,Benaych-Georges2011-nq}, and the random energy model \cite{Derrida1980-xw}. 
    
    The outline of this paper is as follows. We set up the basic notation and questions we ask about gradients and Hessians in detail Sec.~\ref{sec:overall}.  In Sec.~\ref{sec:gradhessana} we introduce a sequence of simplifying assumptions about the structure of gradients, Hessians, logit gradients and logit curvatures that enable us to obtain in the end an extremely simple random model of Hessians and gradients and how they evolve both over training time and weight scale.  We then immediately demonstrate in Sec.~\ref{sec:experiments} that all $4$ striking properties of local geometry of the loss landscape emerge naturally from our simple random model.  Finally, in Sec.~\ref{sec:justification} we give direct evidence that our simplifying theoretical assumptions leading to our random model in Sec.~\ref{sec:gradhessana} are indeed valid in practice, by performing numerical experiments on realistic architectures and datasets.

	\section{Overall framework}
	\label{sec:overall}
	
	Here we describe the local shape of neural loss landscapes, as quantified by their gradient and Hessian, and formulate the main problem we aim to solve: conceptually understanding the emergence of the $4$ striking properties from these two fundamental objects. 
	
	\subsection{Notation and general setup}
	We consider a classification task with $C$ classes. 
	Let $\{\vec{x}^\mu,\vec{y}^\mu\}_{\mu = 1}^N$ denote a dataset of $N$ input-output vectors where the outputs $\vec{y}^{\mu} \in \mathbb{R}^C$ are one-hot vectors, with all components equal to $0$ except a single component ${y}^{\mu}_k=1$ if and only if $k$ is the correct class label for input $x^\mu$.  
	We assume a neural network transforms each input $\vec{x}^\mu$ into a logit vector $\vec{z}^\mu \in \mathbb{R}^C$ through the function $\vec{z}^\mu = \mathcal{F}_{\vec{W}}(x^\mu)$, where $\vec{W} \in \mathbb{R}^D$ denotes a $D$ dimensional vector of trainable neural network parameters. 
	We aim to obtain the aforementioned 4 local properties of the loss landscape as a consequence of a set of simple properties and therefore we do not assume the function $\mathcal{F}_{\vec{W}}$ corresponds to any particular architecture such as a ResNet \cite{he2016deep}, deep convolutional neural network \cite{lecun1989backpropagation}, or a fully-connected neural network. 
	We do assume though that the predicted class probabilities $p^\mu_k$ are obtained from the logits $z^\mu_k$ via the softmax function as
	\begin{equation}
	p^\mu_k = \mathrm{softmax}(\vec{z}^\mu)_k = \frac{\exp{z^\mu_k}}{\sum_{l=1}^C \exp{z^\mu_l} } \, .
	\label{eq:softmax}    
	\end{equation}
	
	We assume network training proceeds by minimizing the widely used cross-entropy loss, which on a particular input-output pair $\{\vec{x}^\mu,\vec{y}^\mu\}$ is given by 
	\begin{equation}
	\mathcal{L}^\mu = - \sum_{k=1}^C y^\mu_k \log \left ( p^\mu_k \right ).
	\label{eq:singleloss}
	\end{equation}
	The average loss over the dataset then yields a loss landscape  $\mathcal{L}$ over the trainable parameter vector $\vec{W}$:
	\begin{equation}
	\label{eq:landscape}
	\mathcal{L}(\vec{W}) = \frac{1}{N} \sum_{\mu=1}^N L^\mu =  - \frac{1}{N} \sum_{\mu=1}^N \sum_{k=1}^C y^\mu_k \log \left ( p^\mu_k \right ) \, .
	\end{equation}
	
	\subsection{The gradient and the Hessian}
	In this work we are interested in two fundamental objects that characterize the local shape of the loss landscape $\mathcal{L}(\vec{W})$, namely its slope, or gradient $\vec{g} \in \mathbb{R}^D$, with components given by
	\begin{equation}
	g_\alpha = \frac{\partial \mathcal{L}}{\partial W_\alpha} \, ,
	\label{eq:graddef}
	\end{equation}
	and its local curvature, defined by the $D \times D$ Hessian matrix $H$, with matrix elements given by
	\begin{equation}
	H_{\alpha \beta} = \frac{\partial^2 \mathcal{L}}{\partial W_\alpha \partial W_\beta} \, .
	\label{eq:hessdef}
	\end{equation}
	Here, $W_\alpha$ is the $\alpha^\mathrm{th}$ trainable parameter, or weight specifying $\mathcal{F}_{\vec{W}}$. 
	Both the gradient and Hessian can vary over weight space $\vec{W}$, and therefore over training time, in nontrivial ways.  
	
	In general, the loss $\mathcal{L}^\mu$ in \eqref{eq:singleloss} depends on the logit vector $\vec{z}^\mu$, which in-turn depends on the weights $\vec{W}$ as $\mathcal{L}^\mu(\vec{z}^\mu(\vec{W}))$.  We can thus obtain explicit expressions for the gradient and Hessian with respect to weights $\vec{W}$ in \eqref{eq:graddef} and \eqref{eq:hessdef} by first computing the gradient and Hessian with respect the logits $\vec{z}$ and then applying the chain-rule.  Due to the particular form of the soft-max function in \eqref{eq:softmax} and cross-entropy loss in \eqref{eq:singleloss}, the gradient of the loss $\mathcal{L}^\mu$ with respect to the logits $\vec{z}^\mu$ is
	\begin{equation}
	\left ( \nabla_z{L^\mu} \right )_k = \frac{\partial L^\mu}{\partial z^\mu_k} =  y^\mu_k - p^\mu_k  \, ,
	\label{eq:gradlosslogit}
	\end{equation}
	and the Hessian of the loss $\mathcal{L}^\mu$ with respect to logits is
	\begin{equation}
	\left ( \nabla^2_z{L^\mu} \right )_{k l} = \frac{\partial^2 L^\mu}{\partial z^\mu_k \partial z^\mu_l} =  p^\mu_k \left ( \delta_{k l} - p^\mu_l \right )  \, .
	\label{eq:probs_crossterm}
	\end{equation}
	
	Then applying the chain rule yields the gradient of $\mathcal{L}$ w.r.t. the weights as
	\begin{equation}
	\label{eq:full_gradient}
	g_\alpha = \frac{1}{N} \sum_{\mu=1}^N \sum_{k=1}^C \left ( y^\mu_k - p^\mu_k \right ) \frac{\partial z^\mu_k}{\partial W_\alpha} \, 
	\end{equation}
	The chain rule also yields the Hessian in \eqref{eq:hessdef}:
	\begin{multline}
	\label{eq:full_Hessian}
	H_{\alpha \beta} = \underbrace{\frac{1}{N}\sum_{\mu=1}^N \sum_{k=1}^C \sum_{l=1}^C \frac{\partial z^\mu_k}{\partial W_\alpha} \left ( \nabla^2_z{L^\mu} \right )_{k l}  \frac{\partial z^\mu_l}{\partial W_\beta}}_{\mathrm{G-term}} +\\+ \underbrace{\frac{1}{N}\sum_{\mu=1}^N \sum_{k=1}^C  \left ( \nabla_z{L^\mu} \right )_k \frac{\partial^2 z^\mu_k}{\partial W_\alpha \partial W_\beta}}_{\mathrm{H-term}}
	\, .
	\end{multline}
	The Hessian consists of a sum of two-terms which have been previously referred to as the \text{G-term} and \text{H-term} \cite{papyan2019measurements}, and we adopt this nomenclature here. 
	
	The basic equations \eqref{eq:gradlosslogit}, \eqref{eq:probs_crossterm},  \eqref{eq:full_gradient} and \eqref{eq:full_Hessian} constitute the starting point of our analysis. They describe the explicit dependence of the gradient and Hessian on a host of quantities: the correct class labels $y^\mu_k$, the predicted class probabilities $p^\mu_k$, the logit gradients $\frac{\partial z^\mu_k}{\partial W_\alpha}$ and logit curvatures
	$\frac{\partial^2 z^\mu_k}{\partial W_\alpha \partial W_\beta}$. 
	It is conceptually unclear how the $4$ striking properties of the local shape of neural network loss landscapes described in Sec. \ref{sec:intro} all emerge naturaly from the explicit expressions in equations \eqref{eq:gradlosslogit}, \eqref{eq:probs_crossterm},  \eqref{eq:full_gradient} and \eqref{eq:full_Hessian}, and moreover, which {\it specific} properties of class labels, probabilities and logits play a key role in their emergence.
	
	\section{Analysis of the gradient and Hessian}
	\label{sec:gradhessana}
	
	In the following subsections, through a sequence of approximations, motivated both by theoretical considerations and empirical observations, we isolate three key features that are sufficient to explain the 4 striking properties: (1) weak logit curvature, (2) clustering of logit gradients, and (3) freezing of class probabilities, both over training time and weight scale.  We discuss each of these features in the next three subsections.  
	
	\subsection{Weakness of logit curvature}
	\label{subsec:logitcurv}
	We first present a combined set of empirical and theoretical considerations which suggest that the \textit{G-term} dominates the \textit{H-term} in \eqref{eq:full_Hessian}, in determining the structure of the top eigenvalues and eigenvectors of neural network Hessians.  First, empirical studies \cite{ghorbani2019investigation,Sagun2017EigenvaluesOT,sagun2017empirical} demonstrate that large neural networks trained on real data with gradient descent have Hessian eigenspectra consisting of a continuous bulk eigenvalue distribution plus a small number of large outlier eigenvalues.  Moreover, some of these studies have shown that the spectrum of the H-term alone is similar to the bulk spectrum of the total Hessian, while the spectrum of the G-term alone is similar to the outlier eigenvalues of the total Hessian.  
	
	This bulk plus outlier spectral structure is extremely well understood in a wide array of simpler random matrix models \cite{Baik2005-rh,Benaych-Georges2011-nq}.  Without delving into mathematical details, a common observation underlying these models is if $H = A + E$ where $A$ is a low rank large $N \times N$ matrix with a small number of nonzero eigenvalues $\lambda^A_i$, while $E$ is a full rank random matrix with a bulk eigenvalue spectrum, then as long as the eigenvalues $\lambda^A_i$ are large relative to the scale of the eigenvalues of $E$, then the spectrum of $H$ will have a bulk plus outlier structure.  In particular, the bulk spectrum of $H$ will look similar to that of $E$, while the outlier eigenvalues $\lambda^H_i$ of $H$ will be close to the eigenvalues $\lambda^A_i$ of $A$.  However, as the scale of $E$ increases, the bulk of $H$ will expand to swallow the outlier eigenvalues of $H$.  An early example of this sudden loss of outlier eigenvalues is known as the BBP phase transition \cite{Baik2005-rh}.         
	
	In this analogy $A$ plays the role of the G-term, while $E$ plays the role of the H-term in \eqref{eq:full_Hessian}.  
	However, what plausible training limits might diminish the scale of the H-term compared to the G-term to ensure the existence of outliers in the full Hessian?  
	Indeed recent work exploring the \textit{Neural Tangent Kernel}
	\cite{jacot2018neural,lee2019wide} assumes that the 
	logits $z^\mu_k$ depend only linearly on the weights $W_\alpha$, 
	which implies that the logit curvatures $\frac{\partial^2 z^\mu_k}{\partial W_\alpha \partial W_\beta}$, and therefore the H-term are {\it identically} zero. 
	More generally, if these logit curvatures are weak over the course of training (which one might expect if the NTK training regime is similar to training regimes used in practice), 
	then one would expect based on analogies to simpler random matrix models, that the outliers of $H$ in \eqref{eq:full_gradient} are well modelled by the G-term alone, as empirically observed previously \cite{papyan2019measurements}.

	Based on these combined empirical and theoretical arguments, we model the Hessian using the \textit{G-term} only:	
	\begin{equation}
	H^\mathrm{model}_{\alpha \beta} = \frac{1}{N}\sum_{\mu=1}^N \sum_{k,l=1}^C  \frac{\partial z^\mu_k}{\partial W_\alpha}p^\mu_k \left ( \delta_{k l} - p^\mu_l \right )  \frac{\partial z^\mu_l}{\partial W_\beta} \, ,
	\label{eq:NTK_Hessian}
	\end{equation}
	where we used \eqref{eq:probs_crossterm} for the Hessian w.r.t. logits.
	\begin{figure*}[ht]
		\centering
		\includegraphics[width=1.0\linewidth]{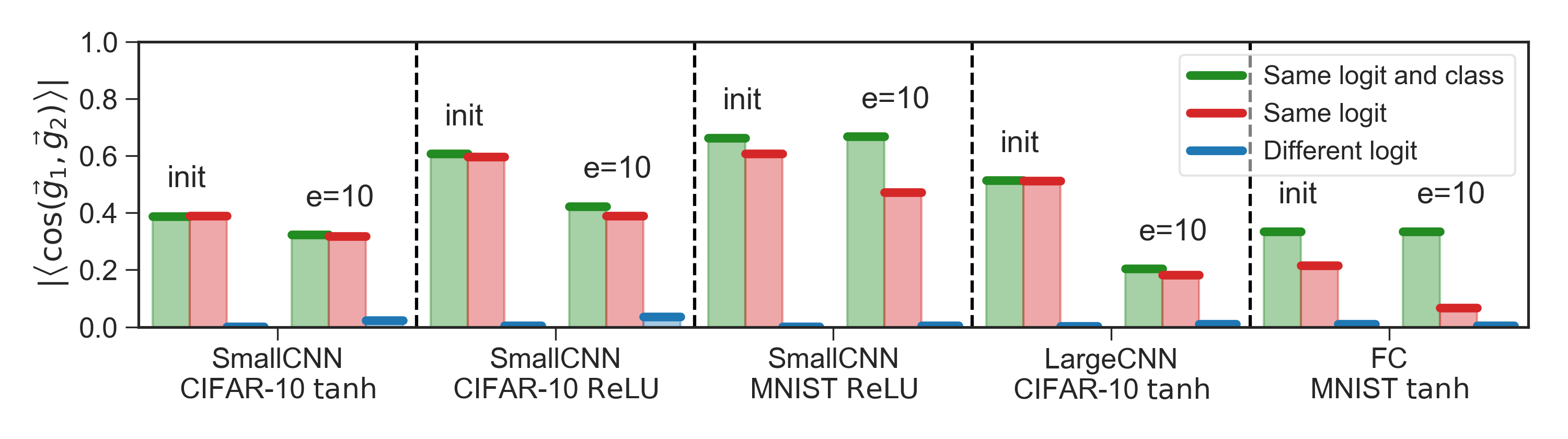}
		\caption{The experimental results on clustering of logit gradients for different datasets, architectures, non-linearities and stages of training. The green bars correspond to $q^{\mathrm{SLSC}}$ in Eq.~\ref{eq:q_SLSC}, the red bars to $q^{\mathrm{SL}}$ in Eq.~\ref{eq:q_SL}, and the blue bars to $q^{\mathrm{DL}}$ in Eq.~\ref{eq:q_DL}. In general, the gradients with respect to weights of logits $k$ will cluster well regardless of the class of the datapoint $\mu$ they were evaluated at. For datapoints of true class $k$, they will cluster slightly better, while gradients of two logits $k \neq l$ will be nearly orthogonal. This is visualized in Fig~\ref{fig:logit_gradient_alignment_figure}.}.
		\label{fig:logit_gradient_alignment}
	\end{figure*}
	\subsection{Clustering of logit gradients}
	\label{sec:model1}
	
	We next examine the logit gradients $\frac{\partial z^\mu_k}{\partial W_\alpha}$, which play a prominent role in both the Hessian (after neglecting logit curvature) in \eqref{eq:NTK_Hessian} and the gradient in \eqref{eq:full_gradient}.  Previous work \cite{fort2019stiffness} noted that gradients of the loss $\mathcal{L}^\mu$ cluster based on the correct class memberships of input examples $\vec{x}^\mu$.  While the loss gradients are not exactly the logit gradients, they are composed of them. Based on our own numerical experiments, we investigated and found strong logit gradient clustering on a range of networks, architectures, non-linearities, and datasets as demonstrated in Figure~\ref{fig:logit_gradient_alignment} and discussed in detail Sec.~\ref{subsec:justifylogitclust}. In particular, we examined three measures of logit gradient similarity.  First, consider
	
	\begin{equation}
	q^{\text{SLSC}} = 
	\frac{1}{C} \sum_{k=1}^C \frac{1}{N_k(N_k-1)} 
	\sum_{\substack{\mu,\nu \in k\\ \mu \neq \nu}}
	\cos \angle \left( \frac{\partial z^\mu_k}{\partial \vec{W}} , \frac{\partial z^\nu_k}{\partial \vec{W}} \right). 
	\label{eq:q_SLSC}
	\end{equation}
	Here SLSC is short for Same-Logit-Same-Class, and $q^{\text{SLSC}}$ measures the average cosine similarity over all pairs of logit gradients corresponding to the same logit component $k$, and all pairs of examples $\mu$ and $\nu$ belonging to the same desired class label $k$.  $N_k$ denotes the number of examples with correct class label $k$.  Alternatively, one could consider  
	\begin{equation}
	q^{\text{SL}} = 
	\frac{1}{C} \sum_{k=1}^C \frac{1}{N(N-1)} 
	\sum_{\mu \neq \nu}
	\cos \angle \left( \frac{\partial z^\mu_k}{\partial \vec{W}} , \frac{\partial z^\nu_k}{\partial \vec{W}} \right). 
	\label{eq:q_SL}
	\end{equation}
	Here SL is short for Same-Logit and $q^{\text{SL}}$ measures the average cosine similarity over all pairs of logit gradients corresponding to the same logit component $k$, and all pairs of examples $\mu \neq \nu$, {\it regardless} of whether the correct class label of examples $\mu$ and $\nu$ is also $k$.  Thus $q^{\text{SLSC}}$ averages over more restricted set of example pairs than does $q^{\text{SL}}$. Finally, consider the null control 
	\begin{equation}
	q^{\text{DL}} = 
	\frac{1}{C(C-1)} \sum_{k \neq l} \frac{1}{N(N-1)} 
	\sum_{\mu \neq \nu}
	\cos \angle \left( \frac{\partial z^\mu_k}{\partial \vec{W}} , \frac{\partial z^\nu_l}{\partial \vec{W}} \right). 
	\label{eq:q_DL}
	\end{equation}
	Here DL is short for Different-Logits and $q^{\text{DL}}$ measures the average cosine similarity for all pairs of different logit components $k\neq l$ and all pairs of examples $\mu \neq \nu$. 
	
	Extensive numerical experiments detailed in Figure~\ref{fig:logit_gradient_alignment} and in Sec.~\ref{subsec:justifylogitclust} demonstrate that both $q^{\text{SLSC}}$ and $q^{\text{SL}}$ are large relative to $q^{\text{DL}}$, implying: (1) logit gradients of logit $k$ cluster together for inputs $\mu$ whose ground truth class is $k$; (2) logit gradients of logit $k$ also cluster 
	\begin{figure}[ht]
		\centering
		\includegraphics[width=0.6\linewidth]{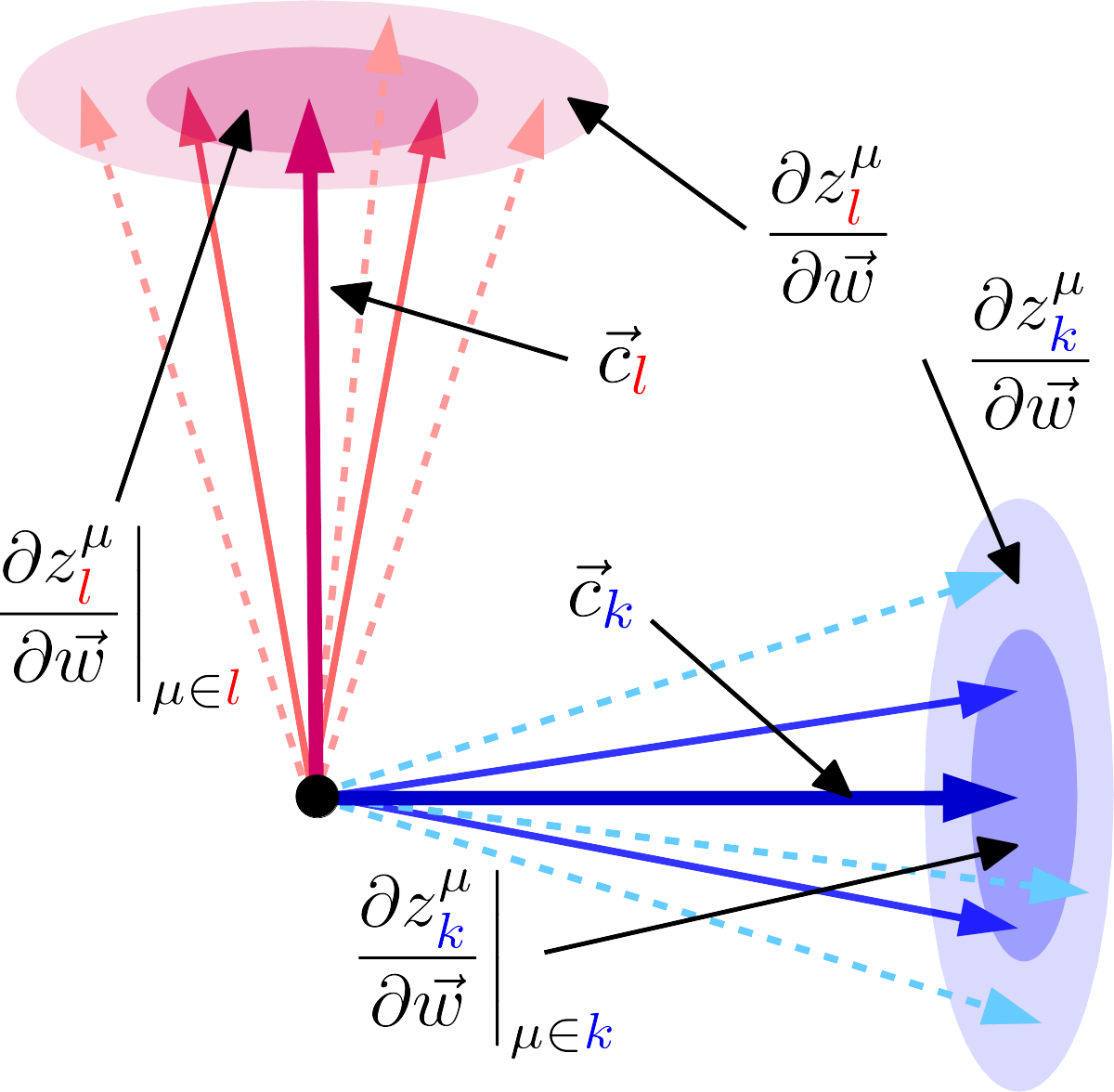}
		\caption{A diagram of logit gradient clustering. The \kth logit gradients cluster based on $k$, regardless of the input datapoint $\mu$. The gradients coming from examples $\mu$ of the class $k$ cluster more tightly, while gradients of different logits $k$ and $l$ are nearly orthogonal.}
		\label{fig:logit_gradient_alignment_figure}
	\end{figure}
	together regardless of the class label of each example $\mu$, although slightly less strongly than when the class label is restricted to $k$; (3) logit gradients of two different logits $k$ and $l$ are essentially orthogonal; (4) such clustering occurs not only at initialization but also after training.
	
	Overall, these results can be viewed schematically as in Figure~\ref{fig:logit_gradient_alignment_figure}.  Indeed, one can decompose the logit gradients as 
	\begin{equation}
	\label{eq:model1_B}
	\frac{\partial z^\mu_k}{\partial W_\alpha} \equiv c_{k \alpha} + E^\mu_{k \alpha} \, ,
	\end{equation}
	where the C vectors $\left \{ \vec{c}_k \in \mathbb{R}^D \right \}_{k=1}^C$ have components 
	\begin{equation}
	c_{k\alpha} = \frac{1}{N_k}\sum_{\mu \in k}^{} \frac{\partial z^\mu_k}{\partial {W_\alpha}} \, , 
	\label{eq:meanlogitgrad}
	\end{equation}
	and $E^\mu_{k \alpha}$ denotes the example specific residuals.  Clustering, in the sense of large $q^{\text{SL}}$, implies the mean logit gradient components $c_{k\alpha}$ are significantly larger than the residual components $E^\mu_{k \alpha}$. In turn the observation of small $q^{\text{DL}}$ implies that mean logit gradient vectors $\vec{c}_k$ and $\vec{c}_l$ are essentially orthogonal.  Both effects are depicted schematically in Fig.~\ref{fig:logit_gradient_alignment_figure}.  Overall, this observation of logit gradient clustering is similar to that noted in \cite{pmlr-v97-papyan19a}, though the explicit numerical modeling and the focus on the $4$ properties in Sec.~\ref{sec:intro} goes beyond it. 
	
	Equations \eqref{eq:NTK_Hessian} and \eqref{eq:model1_B} and  suggest a random matrix approach to modelling the Hessian, as well as a random model for the gradient in \eqref{eq:full_gradient}.  
	The basic idea is to model the mean logit gradients 
	$c_{k\alpha}$, 
	the residuals $E^\mu_{k \alpha}$, 
	and the logits $z^\mu_k$ themselves 
	(which give rise to the class probabilities $p^\mu_k$ through \eqref{eq:softmax}) as {\it independent} random variables. 
	Such a modelling approach neglects correlations between logit gradients and logit values across both examples and weights.  
	However, we will see that this simple modelling assumption is sufficient to produce the 4 striking properties of the local shape of neural loss landscapes described in Sec.~\ref{sec:intro}. 
	
	In this random matrix modelling approach, we simply choose the components $c_{k\alpha}$ to be i.i.d. zero mean Gaussian variables with variance $\sigma^2_c$, while we choose the residuals to be i.i.d. zero mean Gaussians with variance $\sigma^2_E$. With this choice, for high dimensional weight spaces with large $D$, we can realize the logit gradient geometry depicted in Fig.~\ref{fig:logit_gradient_alignment_figure}.  Indeed the mean logit gradient vectors $\vec{c}_k$ are approximately orthogonal, and logit gradients cluster at high $\text{SNR} = \frac{\sigma^2_c}{\sigma^2_E}$ with a clustering value given by $q^{\text{SL}} = \frac{\text{SNR}}{\text{SNR}+1}$.  Finally, inserting the decomposition \eqref{eq:model1_B} into \eqref{eq:NTK_Hessian} and neglecting cross-terms whose average would be negligible at large $N$ due to the assumed independence of the logit gradient residuals $E^\mu_{k\alpha}$ and logits $z^\mu_k$ in our model, yields 
	\begin{multline}
	H^\mathrm{model}_{\alpha \beta} =  \sum_{k,l=1}^C  c_{k\alpha} \left[ \frac{1}{N}\sum_{\mu=1}^N p^\mu_k \left ( \delta_{k l} - p^\mu_l \right )  \right] c_{l\beta} \quad + \\
	\frac{1}{N}\sum_{\mu=1}^N \sum_{k,l=1}^C  E^\mu_{k\alpha} p^\mu_k \left ( \delta_{k l} - p^\mu_l \right )  E^{\mu}_{l\beta} \, .
	\label{eq:clusteredHessian}
	\end{multline}
	This is the sum of a rank $C$ term with a high rank noise term, and the larger the logit clustering $q^{\text{SL}}$, the larger the eigenvalues of the former relative to the latter, yielding $C$ outlier eigenvalues plus a bulk spectrum through the BBP analogy described in Sec.~\ref{subsec:logitcurv}. 
	
	While these choices constitute our random model of logit-gradients, to complete the random model of both the Hessian in \eqref{eq:clusteredHessian} and the gradient in \eqref{eq:full_gradient}, we need to provide a random model for the logits $z^\mu_k$, or equivalently the class probabilities $p^\mu_k$, which we turn to next.
	
	\subsection{Freezing of class probabilities both over training time and weight scale}
	\label{subsec:freezing}
	\begin{figure*}[ht]
	\centering
	\includegraphics[width=1.0\linewidth]{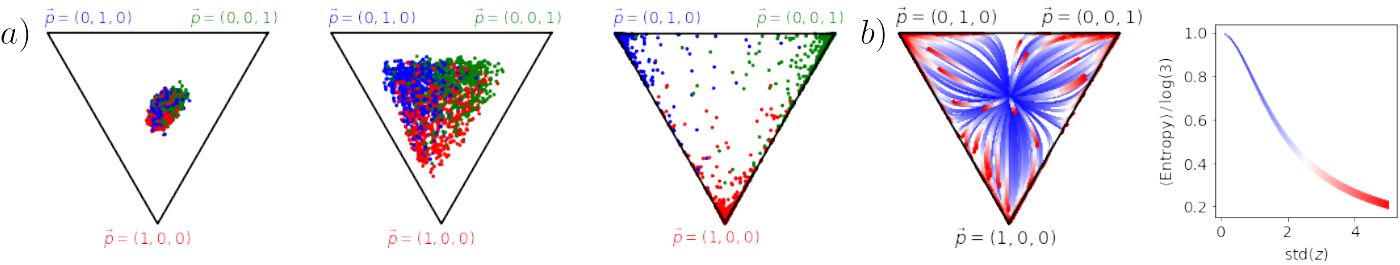}
	\caption{The motion of probabilities in the probability simplex a) during training in a real network, and b) as a function of logit variance $\sigma_z$ in our random model. (a) The distribution of softmax probabilities in the probability simplex for a 3-class subset of CIFAR-10 during an early, middle, and late stage of training a SmallCNN. (b) The motion of probabilities induced by increasing the logit variance $\sigma^2_z$ (blue to red) in our random model and the corresponding decrease in the entropy of the resulting distributions.}
	\label{fig:prob_simplex_3_classes}
	\end{figure*}
	A common observation is that over training time, the predicted softmax class probabilities $p^\mu_k$ evolve from hot, or high entropy distributions near the center of the $C-1$ dimensional probability simplex, to colder, or lower entropy distributions near the corners of the same simplex, where the one-hot vectors $y^\mu_k$ reside. 
	An example is shown in Fig.~\ref{fig:prob_simplex_3_classes} for the case of $C=3$ classes of CIFAR-10.  We can develop a simple random model of this freezing dynamics over training by assuming the logits $z^\mu_k$ themselves are i.i.d. zero mean Gaussian variables with variance $\sigma^2_z$, and further assuming that this variance $\sigma^2_z$ increases over training time.  Direct evidence for the increase in logit variance over training time is presented in Fig.~\ref{fig:logit_variance_and_logit_gradiation_length_and_weight_radius_vs_epoch} and in Sec.~\ref{subsec:justifylogitvar}. 
	
	The random Gaussian distribution of logits $z^\mu_k$ with variance $\sigma^2_z$ in turn yields a random probability vector $\vec{p}^\mu \in \mathbb{R}^C$ for each example $\mu$ through the softmax function in \eqref{eq:softmax}.  This random probability vector is none other than that found in Derrida's famous {\it random energy model} \cite{Derrida1980-xw}, which is a prototypical toy example of a spin glass in physics.  Here the negative logits $-z^\mu_k$ play the role of an energy function over $C$ physical states, the logit variance $\sigma^2_z$ plays the role of an inverse temperature, and $\vec{p}^\mu$ is thought of as a Boltzmann distribution over the $C$ states.  At high temperature (small $\sigma^2_z$), the Boltzmann distribution explores all states roughly equally yielding an entropy $S = - \sum_{k=1}^C p^\mu_k \log_2 p^\mu_k \approx \log_2 C$.  Conversely as the temperature decreases ($\sigma^2_z$ increases), the entropy $S$ decreases, approaching $0$, signifying a frozen state in which $\vec{p}^\mu$ concentrates on one of the $C$ physical states (with the particular chosen state depending on the particular realization of energies $-z^\mu_k$).  Thus this simple i.i.d. Gaussian random model of logits mimics the behavior seen in training simply by increasing $\sigma^2_z$ over training time, yielding the observed freezing of predicted class probabilities (Fig.~\ref{fig:prob_simplex_3_classes}).  
	
	Such a growth in the scale of logits over training is indeed demonstrated in Fig.~\ref{fig:logit_variance_and_logit_gradiation_length_and_weight_radius_vs_epoch} and in Sec.~\ref{subsec:justifylogitvar}, and it could arise naturally as a consequence of an increase in the scale of the weights over training, which has been previously reported  \cite{fort2019large,anonymous2020deep}. We note also that the same freezing of predicted softmax class probabilities could also occur at initialization as one moves radially out in weight space, which would then increase the logit variance $\sigma^2_z$ as well.  Below in Sec. \ref{sec:experiments} we will make use of the assumed feature of freezing of class probabilities both over increasing training times and over increasing weight scales at initialization.

	\section{Deriving loss landscape properties}
	\label{sec:experiments}
	
	We are now in a position to exploit the features and simplifying assumptions made in Sec.~\ref{sec:gradhessana} to provide an exceedingly simple unifying model for the gradient and the Hessian that simultaneously accounts for all 4 striking properties of the neural loss landscape described in Sec.~\ref{sec:intro}.  We first review the essential simplifying assumptions.  First, to understand the top eigenvalues and associated eigenvectors of the Hessian, we assume the logit curvature term in \eqref{eq:full_Hessian} is weak enough to neglect, yielding the model Hessian in \eqref{eq:NTK_Hessian}, which is composed of logit gradients $\partial z^\mu_k / \partial W_\alpha$ and predicted class probabilities $p^\mu_k$. In turn, these quantities could {\it a priori} have complex, interlocking dependencies both over weight space and over training time, leading to potentially complex behavior of the Hessian and its relation to the gradient in \eqref{eq:full_gradient}. 
	
	We instead model these quantities by simply employing a set of independent zero mean Gaussian variables with specific variances that can change over either training or over weight space.  
	We assume the logit gradients decompose as in \eqref{eq:model1_B} with mean logit gradients distributed as $c_{k\alpha} \sim \mathcal{N}(0,\sigma^2_c)$ 
	and residuals distributed as  $E^\mu_{k\alpha} \sim \mathcal{N}(0,\sigma^2_E)$. 
	Additionally, we assume the logits $z^\mu_k$ themselves are distributed as $z^\mu_{k\alpha} \sim \mathcal{N}(0,\sigma^2_z)$.  
	As we see next, inserting this collection of i.i.d. Gaussian random variables into the expressions for the softmax in \eqref{eq:softmax}, the gradient in \eqref{eq:full_gradient}, and the Hessian model in \eqref{eq:clusteredHessian}, for various choices of $\sigma^2_c$, $\sigma^2_E$ and $\sigma^2_z$, yields a simple unified model sufficient to account for the 4 striking observations in Sec.~\ref{sec:intro}.
	
\begin{figure}[ht]
		\centering
        \includegraphics[width=0.48\linewidth]{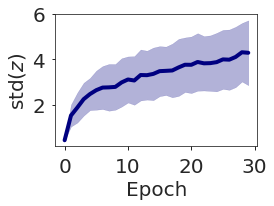}
        \includegraphics[width=0.48\linewidth]{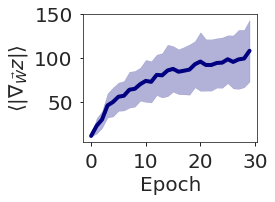}\\
		\includegraphics[width=0.48\linewidth]{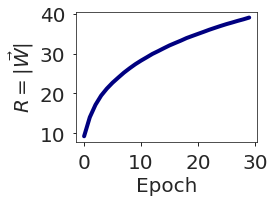}
		\includegraphics[width=0.48\linewidth]{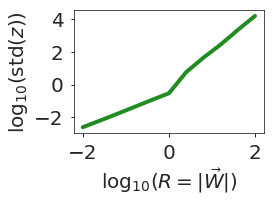}
		\caption{The evolution of logit variance, logit gradient length, and weight space radius  with training time. The top left panel shows that the logit variance across classes, averaged over examples, grows with training time. The top right panel shows that logit gradient lengths grow with training time. The bottom left panel shows the weight norm grows with training time. All 3 experiments were conducted with a SmallCNN on CIFAR-10. The bottom right panel shows the logit variance grows as one moves radially out in weight space, at random initialization, with no training involved, again in a SmallCNN.}
		\label{fig:logit_variance_and_logit_gradiation_length_and_weight_radius_vs_epoch}
	\end{figure}

\begin{figure}[ht]
			\centering
			\includegraphics[width=1.0\linewidth]{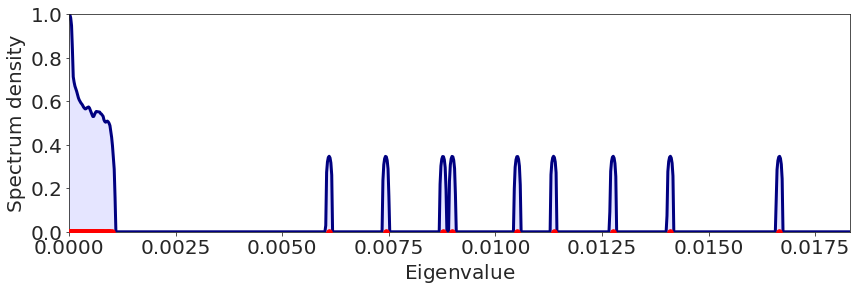}
			\caption{The Hessian eigenspectrum in our random model. Due to logit-clustering it exhibits a bulk + $C-1$ outliers. To obtain $C$ outliers, we can use mean logit gradients $\vec{c}_k$ whose lengths vary with $k$ (data not shown).}
			\label{fig:Hessian_spectrum_sorted}
		\end{figure}

\begin{figure}[ht]
			\centering
			\includegraphics[width=1.0\linewidth]{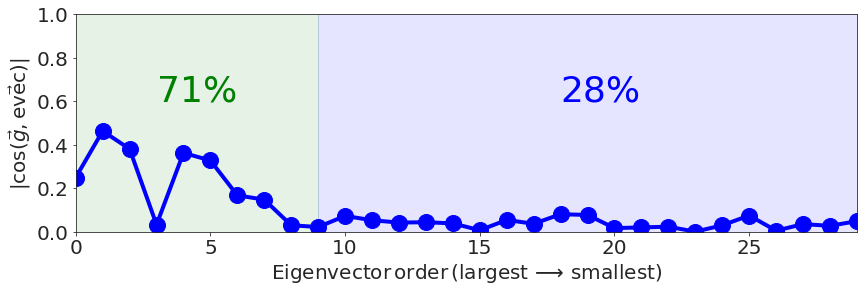}
			\caption{The overlap between Hessian eigenvectors and gradients in our random model. Blue dots denote cosine angles between the gradient and the sorted eigenvectors of the Hessian. The bulk (71\% in this particular case) of the total gradient power lies in the top 10 eigenvectors (out of $D=1000$) of the Hessian.}
			\label{fig:evec_grad_coses}
		\end{figure}

\begin{figure}[ht]
			\centering
			\includegraphics[width=1.0\linewidth]{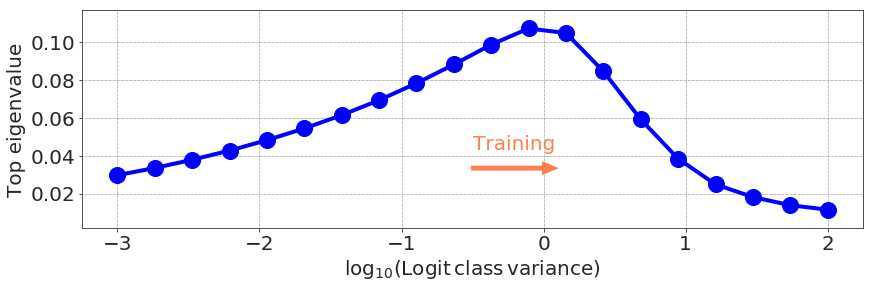}
			\caption{The top eigenvalue of the Hessian in our random model as a function of the logit standard deviation $\sigma_z$ ($\propto$ training time as demonstrated in Fig.~\ref{fig:logit_variance_and_logit_gradiation_length_and_weight_radius_vs_epoch}). We also model logit gradient growth over training by monotonically increasing $\sigma_c$ while keeping $\sigma_c/\sigma_E$ constant.}
			\label{fig:top_eval}
		\end{figure}

\begin{figure}[ht]
			\centering
			\includegraphics[width=1.0\linewidth]{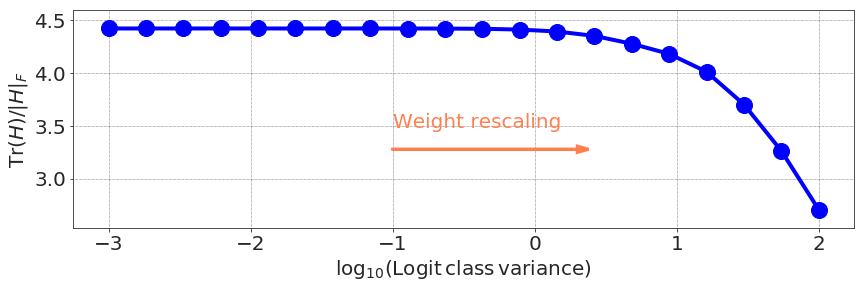}
			\caption{The $\mathrm{Trace}(H)/||H||$ as a function of the logit standard deviation $\sigma_z$ ($\propto$ training time as show in Fig.~\ref{fig:logit_variance_and_logit_gradiation_length_and_weight_radius_vs_epoch}). This transition is equivalent to what was seen for CNNs in \cite{fort2018goldilocks}.
			}
			\label{fig:trace}
		\end{figure}

    The results of our model, shown in Fig.~\ref{fig:Hessian_spectrum_sorted}, \ref{fig:evec_grad_coses}, \ref{fig:top_eval} and \ref{fig:trace}, were obtained using $N=300$, $C=10$ and $D=1000$. The logit standard deviation was $\sigma_z = 15$, leading to the average highest predicted probability of $p=0.94$. The logit gradient noise scale was $\sigma_E = 0.7 / \sqrt{D}$ and the mean logit gradient scale was $\sigma_c = 1.0 / \sqrt{D}$, leading to a same-logit clustering value of $q^{SL} = 0.67$ that matches our observations in Fig.~\ref{fig:logit_gradient_alignment}. We assigned class labels $y^\mu_k$ to random probability vectors $p^\mu_k$ so as to obtain a simulated accuracy of $0.95$.  For the experiments in Figures \ref{fig:top_eval} and \ref{fig:trace}, we swept through a range of logit standard deviations $\sigma_z$ from $10^{-3}$ to $10^{2}$, while also monotonically increasing $\sigma_c$ as observed in real neural networks in Fig.~\ref{fig:logit_variance_and_logit_gradiation_length_and_weight_radius_vs_epoch}. We now demonstrate that the $4$ properties emerge from our model.

		\paragraph{(1) Hessian spectrum = bulk + outliers.} Our random model in Fig.~\ref{fig:Hessian_spectrum_sorted} clearly exhibits this property, consistent with that observed in much more complex neural networks (e.g. \cite{ghorbani2019investigation}). The outlier emergence is a direct consequence of high logit-clustering (large $q^{SL}$), which ensures that rank $C$ term dominates the high rank noise term in \eqref{eq:clusteredHessian}. This dominance yields the outliers through the BBP phase transition mechanism described in Sec.~\ref{subsec:logitcurv}.  
		\paragraph{(2) Gradient confinement to principal Hessian subspace.} Figure~\ref{fig:evec_grad_coses} shows the cosine angle between the gradient and the Hessian eigenvectors in our random model. The majority of the gradient power lies within the subspace spanned by the top few eigenvectors of the Hessian, consistent with observations in real neural networks \cite{gurari2018gradient}.  This occurs because the large mean logit-gradients $c_{k\alpha}$ contribute both to the gradient in \eqref{eq:full_gradient} and principal Hessian eigenspace in \eqref{eq:clusteredHessian}. 
		\paragraph{(3) Non-monotonic evolution of the top Hessian eigenvalue with training time.} Equating training time with a growth of logit variance $\sigma^2_z$ and a simultaneous growth of $\sigma^2_c$ while keeping $\sigma^2_c/\sigma^2_E$ constant, our random model exhibits eigenvalue growth then shrinkage, as shown in Figure~\ref{fig:top_eval} and consistent with observations on large CNNs trained on realistic datasets \cite{jastrzebski2018relation}.  This non-monotonicity arises from a competition between the shrinkage, due to freezing probabilities $p^\mu_k$ with increasing $\sigma^2_z$, of the eigenvalues of the $C$ by $C$ matrix with components $P_{kl} = \frac{1}{N}\sum_{\mu=1}^N p^\mu_k \left ( \delta_{k l} - p^\mu_l \right )$ in \eqref{eq:clusteredHessian}, and the growth of the mean logit gradients $c_{k\alpha}$ in \eqref{eq:clusteredHessian} due to increasing $\sigma^2_c$. 
		\paragraph{(4) The Golidlocks zone: $\mathrm{Trace}(H)/||H||$ is large (small) for small (large) weight scales.} Equating increasing weight scale with increasing logit scale $\sigma^2_z$, our random model exhibits this property, as shown in Fig.~\ref{fig:trace}, and consistent with observations in CNNs \cite{fort2018goldilocks}. To replicate the experiments in \cite{fort2018goldilocks}, we project our Hessian to a random $d=10$ dimensional hyperplane (data not shown) and verified that the behavior we observe is also numerically correct. This decrease in $\mathrm{Trace}(H)/||H||$ (which is approximately invariant to overall mean logit gradient scale $\sigma^2_c$) is primarily a consequence of freezing of probabilities $p^\mu_k$ with increasing $\sigma^2_z$. 

	\section{Justifying modelling assumptions}
	\label{sec:justification}
	
	Our derivation of the 4 properties of the local shape of the neural loss landscape in Sec.~\ref{sec:experiments} relied on several modelling assumptions in a simple, unified random model detailed in Sec.~\ref{sec:gradhessana}.  
	These assumptions include: (1) neglecting logit curvature (introduced and justified in Sec.~\ref{subsec:logitcurv}), 
	(2) logit gradient clustering (introduced in Sec.~\ref{sec:model1} and justified in Sec.~\ref{subsec:justifylogitclust} below), 
	and (3) increases in logit variance both over training time and weight scale, to yield freezing of class probabilities (introduced in Sec.~\ref{subsec:freezing} and justified in Sec.~\ref{subsec:justifylogitvar} below). 
	\subsection{Logit gradient clustering}
	\label{subsec:justifylogitclust}
	Fig.~\ref{fig:logit_gradient_alignment} demonstrates, as hypothesized in Fig.~\ref{fig:logit_gradient_alignment_figure}, that logit gradients do indeed cluster together within the same logit class, and that they are essentially orthogonal between logit classes. We observed this with fully-connected and convolutional networks, with $\mathrm{ReLU}$ and $\tanh$ non-linearites, at different stages of training (including initialization), and on different datasets.  We note that these measurements are related, but complimentary to the concept of \textit{stiffness} in \cite{fort2019stiffness}.
	\subsection{Logit variance dependence}
	\label{subsec:justifylogitvar}
	Fig.~\ref{fig:logit_variance_and_logit_gradiation_length_and_weight_radius_vs_epoch} demonstrates $4$ empirical facts observed in actual CNNs trained on CIFAR-10 or at initialization: (1) logit variance across classes, averaged over examples, grows with training time; (2)  logit gradient lengths grow with training time; (3) the weight norm grows with training time; (4) logit variance grows with weight scale at random initialization.  These four facts justify modelling assumptions used in our random model of Hessians and gradients: (1) we can model training time by increasing $\sigma^2_z$ corresponding to increasing logit variances in the model; while simultaneously  (2) also increasing $\sigma^2_c$ corresponding to increasing mean logit gradients in the model; (3) we can model increases in weight scale at random initialization by increasing $\sigma^2_z$. We note the connection between training epoch and the weight scale has also been established in \cite{fort2019large, anonymous2020deep}. 
	\section{Discussion}
	Overall, we have shown that four non-intuitive, surprising, and seemingly unrelated  properties of the local geometry of the neural loss landscape can all arise naturally in an exceedingly simple random model of Hessians and gradients and how they vary both over training time and weight scale.  Remarkably, we do not need to make any explicit reference to highly specialized structure in either the data,  the neural architecture, or potential biases in regions explored by gradient descent. Instead the key general properties we required were: (1) weakness of logit curvature;  (2) clustering of logit gradients as depicted schematically in Fig.~\ref{fig:logit_gradient_alignment_figure} and justified in Fig.~\ref{fig:logit_gradient_alignment}; (3) growth of logit variances with training time and weight scale (justified in Fig.~\ref{fig:logit_variance_and_logit_gradiation_length_and_weight_radius_vs_epoch}) which leads to freezing of softmax output distributions as shown in Fig.~\ref{fig:prob_simplex_3_classes}.  Overall, the isolation of these key features provides a simple, unified random model which explains how $4$ surprising properties described in Sec.~\ref{sec:intro} might naturally emerge in a wide range of classification problems.

	\subsubsection*{Acknowledgments}
	We would like to thank Yasaman Bahri and Ben Adlam from Google Brain and Stanislaw Jastrzebski from NYU for useful discussions.
	
	\bibliography{main}
	\bibliographystyle{unsrtnat}

\end{document}